\documentclass[10pt,twocolumn,letterpaper,dvipsnames]{article}

\usepackage{cvpr}              
\makeatletter
\@namedef{ver@everyshi.sty}{}
\makeatother
\usepackage[]{etoolbox}
\usepackage{silence}
\makeatletter
\robustify\@latex@warning@no@line
\makeatother
\usepackage{authblk}
\makeatletter
\renewcommand\AB@affilsepx{ --- \protect\Affilfont}
\makeatother

\usepackage{graphicx}
\usepackage{dblfloatfix} 
\usepackage{amsmath}
\usepackage{amssymb}
\usepackage{booktabs}
\usepackage{sidecap} 
\usepackage{paralist}
\usepackage{multirow}
\usepackage{threeparttable} 
\usepackage{collcell} 
\usepackage[style=ieee, backend=biber,maxnames=6,backref=true]{biblatex} 
\DeclareSourcemap{
  \maps[datatype=bibtex]{
    \map[overwrite]{
      \step[fieldsource=doi, final]
      \step[fieldset=url, null]
      \step[fieldset=eprint, null]
    }  
  }
}
\usepackage[colorlinks]{hyperref}

\usepackage[lighttt]{lmodern} 
\usepackage{listings}
\input{listings-python.prf}
\usepackage[accsupp]{axessibility} 

\newcommand{\nhidden}{10}  
\newcommand{\hlsfml}{\texttt{hls4ml}\xspace}
\newcommand{\QKeras}{\textsc{QKeras}\xspace}

\usepackage{xcolor} 

\usepackage[draft]{pdfcomment} 
\usepackage[hyperfirst=false,acronym,sort=none,shortcuts,nopostdot,style=super,nonumberlist,toc,nogroupskip]{glossaries}

\glsdisablehyper 

\newcommand{\accolor}[1]{\textcolor{Black}{#1}}

\newacronymstyle{myacro}
{%
  \GlsUseAcrEntryDispStyle{long-short}%
}%
{%
  \GlsUseAcrStyleDefs{long-short}%
}

\setacronymstyle{myacro}

\newcommand*{\tip}[1]{
    \ifglsused{#1}{
      {\pdftooltip{\accolor{\glsentryshort{#1}}}{\glsentrydesc{#1}}}%
    }{%
      \gls{#1}
    }%
}%

\newcommand*{\tips}[1]{
    \ifglsused{#1}{
      {\pdftooltip{\accolor{\glsentryshortpl{#1}}}{\glsentrydescplural{#1}}}%
    }{%
      \glspl{#1}
    }%
}%

\newcommand*{\tipshort}[1]{
      {\pdftooltip{\accolor{\glsentryshort{#1}}}{\glsentrydesc{#1}}}%
}%

\newacronym[description={False Negative Rate; signal that is incorrectly classified as noise}]{fnr}{FNR}{False Negative Rate}
\newacronym[description={False Negative; signal event that is incorrectly classified as noise event}]{fn}{FN}{False Negative}
\newacronym[description={False Positive Rate; noise that is incorrectly classified as signal}]{fpr}{FPR}{False Positive Rate}
\newacronym[description={False Positive; noise event that is incorrectly classified as signal event}]{fp}{FP}{False Positive}
\newacronym[description={Guided Event Filtering: Joint Filtering of Intensity Images and Neuromorphic Events}]{gef}{GEF}{Guided Event Flow}
\newacronym[description={Inter Spike Interval (nomenclature from neuroscience)}]{isi}{ISI}{Inter Spike Interval}
\newacronym[description={MAC (Multiply-Accumulate) is the basic operation of signal processing and artificial neural networks. One MAC is 2 Op.}]{mac}{MAC}{Multiply-Accumulate}
\newacronym[description={Multipurpose block random access memory module in FPGA}]{bram}{BRAM}{Block RAM}
\newacronym[description={Register Transfer Logic intermediate form, consisting of combinational and synchronous register logic cells}]{rtl}{RTL}{Register Transfer Logic}
\newacronym[description={Single Threshold Metric; a measure of the ROC TPR/FPR tradeoff at one discrimination threshold}]{stm}{STM}{Single Threshold Metric}
\newacronym[description={Surface of Active Event; image of latest event timestamps, same as Timestamp Image}]{sae}{SAE}{Surface of Active Events}
\newacronym[description={System on Chip; FPGA with embedded programmable processor}]{soc}{SoC}{System on Chip}
\newacronym[description={Time Surface; image of age of events relative to a particular event}]{ts}{TS}{Time Surface}
\newacronym[description={Timestamp Image; 2D image of latest event timstamps, similar to Surface of Active Events}]{ti}{TI}{Timestamp Image}
\newacronym[description={Timestamp+Polarity Image; 2D image of latest event timstamps and +/- brightness change polarites}]{tpi}{TPI}{Timestamp+Polarity Image}
\newacronym[description={True Negative Rate; noise that is correctly classified as noise}]{tnr}{TNR}{True Negative Rate}
\newacronym[description={True Negative; noise that is correctly classified as noise}]{tn}{TN}{True Negative}
\newacronym[description={True Positive Rate; signal that is correctly classified as signal}]{tpr}{TPR}{True Positive Rate}
\newacronym[description={True Positive; signal event that is correctly classified as signal}]{tp}{TP}{True Positive}
\newacronym[longplural={Convolutional Neural Networks}]{cnn}{CNN}{Convolutional Neural Network}
\newacronym[longplural={First In First Out memories}]{fifo}{FIFO}{First In First Out memory}
\newacronym{adc}{ADC}{Analog to Digital Converter}
\newacronym{aer}{AER}{Address Event Protocol}
\newacronym{aps}{APS}{Active Pixel Sensor}
\newacronym{asic}{ASIC}{Application Specific Integrated Circuit}
\newacronym{auc}{AUC}{Area Under the Curve}
\newacronym{baf}{BAF}{Background Activity Filter}
\newacronym{ba}{BA}{Background Activity}
\newacronym{bmof}{BMOF}{Block Matching Optical Flow}
\newacronym{bm}{BM}{Block Matching}
\newacronym{bp}{BP}{Back Propagation}
\newacronym{cfa}{CFA}{Color Filter Array}
\newacronym{cf}{CF}{Complementary Filter}
\newacronym{cg}{CG}{Convolutional Gated Recurrent Unit Network}
\newacronym{cis}{CIS}{CMOS Image Sensor}
\newacronym{cmae}{CMAE}{Cross-Modality Attention Enhancement}
\newacronym{contrastmaximization}{CM}{Contrast Maximization}
\newacronym{cots}{COTS}{Commodity Off-The-Shelf}
\newacronym{cpu}{CPU}{Central Processing Unit}
\newacronym{cv}{CV}{Computer Vision}
\newacronym{davis}{DAVIS}{Dynamic and Active pixel Vision Sensor}
\newacronym{dba}{DBA}{Dynamic Background Activity noise filtering algorithm}
\newacronym{dnn}{DNN}{Deep Neural Network}
\newacronym{dof}{DOF}{Degree of Freedom}
\newacronym{dolp}{DoLP}{Degree of Linear Polarization}
\newacronym{dram}{DRAM}{Dynamic RAM}
\newacronym{drcn}{DRCN}{Deep Recurrent Convolutional Network}
\newacronym{dr}{DR}{Dynamic Range}
\newacronym{dsp}{DSP}{Digital Signal Processing unit}
\newacronym{dvs}{DVS}{Dynamic Vision Sensor}
\newacronym{dwf}{DWF}{Double Window Filter}
\newacronym{e2pd}{E2PD}{Events to Polarization Dataset}
\newacronym{e2p}{E2P}{Events to Polarization}
\newacronym{e2mlp}{E2MLP}{Events to MLP Input}
\newacronym{edflow}{EDFLOW}{Event-driven Optical Flow}
\newacronym{edncnn}{EDnCNN}{Event Denoising CNN}
\newacronym{edp}{EDP}{Event Denoising Precision}
\newacronym{efast}{EFAST}{Event-Based time surface FAST}
\newacronym{epm}{EPM}{Event Probability Mask}
\newacronym{fast}{FAST}{Features from Accelerated Segment Test}
\newacronym{feast}{FEAST}{Feature Extraction with Adaptive Selection Thresholds }
\newacronym{flipflop}{FF}{Flip-Flop}
\newacronym{ff}{FF}{Flip-Flop}
\newacronym{fom}{FOM}{Figure of Merit}
\newacronym{fpga}{FPGA}{Field Programmable Gate Array}
\newacronym{fpn}{FPN}{Fixed Pattern Noise}
\newacronym{fps}{FPS}{Frames Per Second}
\newacronym{fsae}{FSAE}{Filtered Surface of Active Events}
\newacronym{fwf}{FWF}{Fixed Window Filter}
\newacronym{gpu}{GPU}{Graphics Processing Unit}
\newacronym{gt}{GT}{Ground Truth}
\newacronym{hdl}{HDL}{Hardware Description Language}
\newacronym{hdr}{HDR}{high dynamic range}
\newacronym{hls}{HLS}{High Level Synthesis}
\newacronym{icm}{ICM}{Iterated Conditional Modes}
\newacronym{id}{ID}{Index Decay}
\newacronym{iir}{IIR}{Infinite Impulse Response}
\newacronym{imu}{IMU}{Inertial Measurement Unit}
\newacronym{inceptiveevent}{IE}{Inceptive Event}
\newacronym{iot}{IoT}{Internet of Things}
\newacronym{ip}{IP}{Intellectual Property}
\newacronym{its}{ITS}{Invariant Time Surface}
\newacronym{knn}{KNN}{$K$-Nearest-Neighbor clustering}
\newacronym{ldsi}{LDSI}{Less Data Same Information}
\newacronym{li}{LI}{Leaky Integrator}
\newacronym{lk}{LK}{Lucas-Kanade}
\newacronym{lpips}{LPIPS}{Learned Perceptual Image Patch Similarity}
\newacronym{lut}{LUT}{LookUp Table}
\newacronym{mlpf}{MLPF}{MultiLayer Perceptron denoising Filter}
\newacronym{mlp}{MLP}{Multilayer Perceptron}
\newacronym{ml}{ML}{Machine Learning}
\newacronym{mpeg}{MPEG}{Motion Picture Experts Group}
\newacronym{mse}{MSE}{Mean Squared Error}
\newacronym{na}{NA}{Numerical Aperture}
\newacronym{nnb}{NNb}{Nearest Neighbor}
\newacronym{of}{OF}{Optical Flow}
\newacronym{onf}{ONF}{Order(N) Filter}
\newacronym{pcb}{PCB}{Printed Circuit Board}
\newacronym{pdavis}{PDAVIS}{Polarization Dynamic and Active pixel VIsion Sensor}
\newacronym{pd}{PD}{photodiode}
\newacronym{pe}{PE}{Processing Element}
\newacronym{pfa}{PFA}{Polarization Filter Array}
\newacronym{pl}{PL}{programmable Logic}
\newacronym{por}{POR}{Positive Output Ratio}
\newacronym{prm}{PRM}{Pixel Rendering Module}
\newacronym{ps}{PS}{Processing System}
\newacronym{pugm}{PUGM}{Probabilistic Undirected Graph Model}
\newacronym{qwp}{QWP}{Quarter Wave Plate}
\newacronym{ram}{RAM}{Random Access Memory}
\newacronym{ransac}{RANSAC}{Random Sample and Consensus}
\newacronym{ratp}{RATP}{Recursive Adaptive Temporal Pooling}
\newacronym{rb}{RB}{Residual Block}
\newacronym{relu}{ReLU}{Rectified Linear Unit}
\newacronym{roc}{ROC}{Receiver Operating Characteristic}
\newacronym{roi}{ROI}{Region of Interest}
\newacronym{ros}{ROS}{Robot Operating System}
\newacronym{rpmd}{RPMD}{Relative Plausibility Measure of Denoising}
\newacronym{rpm}{RPM}{Revolutions per Minute}
\newacronym{rppp}{RPPP}{Rich Polarization Pattern Perception}
\newacronym{sad}{SAD}{Sum of Absolute Differences}
\newacronym{sm}{SM}{Supplementary Material}
\newacronym{snr}{SNR}{Signal to Noise Ratio}
\newacronym{soa}{SOA}{state of the art}
\newacronym{sram}{SRAM}{Static RAM}
\newacronym{stcf}{STCF}{SpatioTemporal Correlation Filter}
\newacronym{tda}{TDA}{Time Decay Adapted}
\newacronym{td}{TD}{Time Decay}
\newacronym{timsl}{TS}{time slice}
\newacronym{usb}{USB}{Universal Serial Bus}
\newacronym{vga}{VGA}{Video Graphics Adaptor}
\newacronym{vhdl}{VHDL}{Very High-Speed Integrated Circuit Hardware Description Language}
\newacronym{zoh}{ZOH}{Zero-Order Hold}

\usepackage{orcidlink} 

\usepackage{paralist}

\usepackage{makecell}

\usepackage{multirow}

\usepackage[capitalize]{cleveref} 
\crefname{section}{Sec.}{Secs.}
\Crefname{section}{Sec.}{Secs.}
\Crefname{table}{Table}{Tables}
\crefname{table}{Table}{Tables}
\Crefname{figure}{Fig.}{Figs.}
\crefname{figure}{Fig.}{Figs.}

\def\cvprPaperID{14}
\def\confName{CVPR}
\def\confYear{2023}

\title{Within-Camera Multilayer Perceptron DVS Denoising}

\author[2,1,$\dagger$]{A. Rios-Navarro, \orcidlink{0000-0003-4163-8484}}
\author[3,1$\dagger$]{S. Guo, \orcidlink{0000-0002-3308-9123}}
\author[4,$\dagger$]{G Abarajithan \orcidlink{0000-0001-9768-5349}} 
\author[4,$\dagger$]{K. Vijayakumar}
\author[2,*]{A. Linares-Barranco, \orcidlink{0000-0002-6056-740X}}
\author[4,*]{T. Aarrestad, \orcidlink{0000-0002-7671-243X}}
\author[4,1,*]{R. Kastner, \orcidlink{0000-0001-9062-5570}}
\author[1,*]{T. Delbruck,\orcidlink{0000-0001-5479-1141}}

\affil[1]{\footnotesize{Sensors Group, Institute of Neuroinformatics, Univ. of Zurich and ETH Zurich, Switzerland}}
\affil[2]{\footnotesize{Robotic and Tech of Computers group, SCORE lab, ETSI-EPS, Univ. of Seville (USE), Spain}}
\affil[3]{\footnotesize{College of Electronic Engineering, National University of Defense Technology (NUDT), China}}
\affil[4]{\footnotesize{Inst. of Particle Physics and Astrophysics, ETH Zurich (ETH), Switzerland}}
\affil[5]{\footnotesize{Univ. of California, San Diego (UCSD), USA}}
\affil[$\dagger$]{\footnotesize{These authors contributed equally}}
\affil[*]{\footnotesize{Contact authors emails: tobi@ini.uzh.ch, arios@us.es, alinares@atc.us.es, thea.aarrestad@cern.ch, }}

\begin{document}

\twocolumn[{
\renewcommand\twocolumn[1][]{#1}
\vspace{-2cm}
\begin{center}
Accepted to 2023 CVPRW on Event-Based Vision\\
\url{https://tub-rip.github.io/eventvision2023/}
\smallskip
\hrule 
\end{center}
\maketitle
\begin{center}
    \vspace{-6mm}
	\centering
	\captionsetup{type=figure}
	\includegraphics[width=\textwidth]{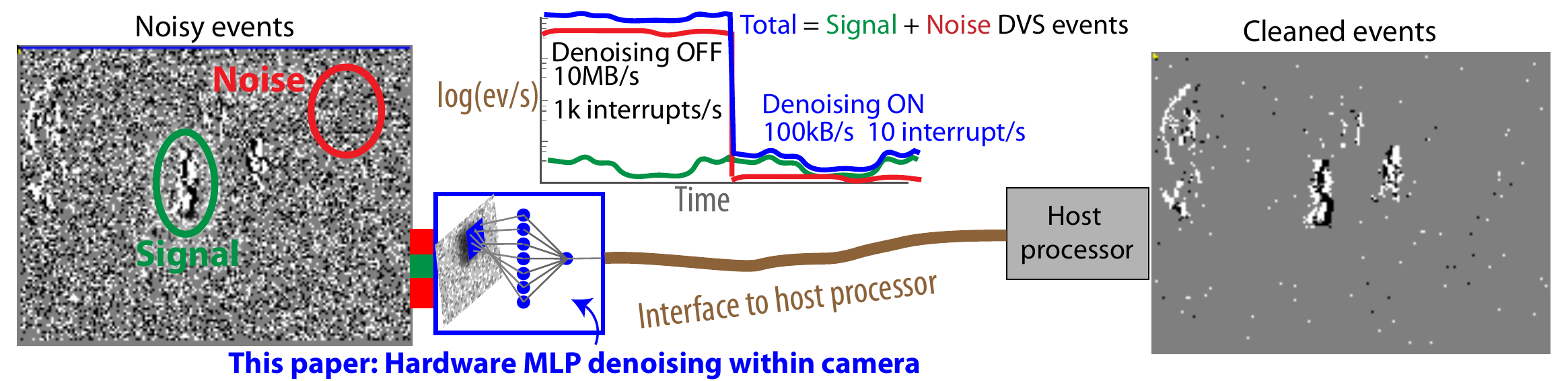}\vspace{-2mm}
    \captionof{figure}{\textbf{Advantages of in-camera denoising}: 
    The \tip{mlpf} reduces the background noise rate in this surveillance scene by a factor of more than 100X while only blocking about 25\% of the true signal events created by the moving people. 
    The \tip{mlpf} accurately and quickly discriminates signal and noise events.
    Building the filter using a hardware-accelerated neural network implemented alongside the camera's logic circuits dramatically reduces the host processing requirements. In low-light scenes where event camera noise increases, the data rate is reduced from 10MB/s to 100kB/s. 
    With a USB buffer size of 10k events, the host processor interrupt rate reduces from 1\,kHz to less than 10\,Hz, 
    allowing the processor to mostly sleep during idle periods.}
	\label{fig:teaser}
	\vspace{-1mm}
\end{center}
}]

\begin{abstract}
In-camera event denoising reduces the data rate of event cameras by filtering out noise at the source. A lightweight multilayer perceptron denoising filter (MLPF) provides state-of-the-art low-cost denoising accuracy. It processes a small neighborhood of pixels from the timestamp image around each event to discriminate signal and noise events. This paper proposes two digital logic implementations of the MLPF denoiser and quantifies their resource cost,  power, and latency. The hardware MLPF quantizes the weights and hidden unit activations to 4 bits and has about 1k weights with about 40\% sparsity. The Area-Under-Curve Receiver Operating Characteristic accuracy is nearly indistinguishable from that of the floating point network. The FPGA MLPF processes each event in 10 clock cycles.  In FPGA, it uses 3.5k flip flops and 11.5k LUTs. Our ASIC implementation in 65nm digital technology for a $346\times 260$ pixel camera occupies an area of 4.3mm$^2$ and consumes 4nJ of energy per event at event rates up to 25MHz. The MLPF can be easily integrated into an event camera using an FPGA or as an ASIC directly on the camera chip or in the same package.
This denoising could dramatically reduce the energy consumed by the communication and host processor and open new areas of always-on event camera application under scavenged and battery power.


\end{abstract}
Code:
\url{https://github.com/SensorsINI/dnd\_hls}

\section{Introduction}
\label{sec:introduction}
\renewcommand*{\thefootnote}{\fnsymbol{footnote}}
Event cameras~\cite{Gallego2020-survey-paper} based on the \tip{dvs}\footnote{The arxiv preprint version of this paper includes links and uses PDF tooltip popups for acronyms that some PDF viewers (Adobe and SumatraPDF version $\le$3.00) can show.} pixel~\cite{Lichtsteiner2008-dvs} are useful for vision problems that require high dynamic range and face the fundamental trade-off between latency and power in frame-based cameras~\cite{Liu2019-eds-ieee-sig-proc}. \tip{dvs} pixels produce events that signal the change in brightness but also produce \tip{ba} noise, particularly under low light conditions, where noise rates can increase by a factor of 100~\cite{hu2021v2e,Guo2022-mlpf-denoising-pami,Finateu2020-isscc-2020-sony,Graca2021-iisw-unraveling-dvs-noise}. 
\renewcommand*{\thefootnote}{\arabic{footnote}}

Denoising algorithms aim to filter out the noise without removing any signal events.
Denoising can be performed on the host computer,
but then the camera must transmit all the raw signal and noise events, consuming more power in the communication bus (e.g. USB) and on the host processor, which must remain awake all the time to remove the noise.

\cref{fig:teaser} shows how denoising within the camera can dramatically reduce the energy consumed by the communication and host processor. Without denoising, the host processor must constantly remain awake and busy even in the absence of any real signal events. Accurate denoising within the camera preserves most signal events but filters out nearly all the noise events. Thus during quiescent periods where the raw camera output is dominated by noise events, the host processor interrupt rate can be reduced by 2 orders of magnitude, allowing it to sleep most of the time and thus open new areas of always-on event camera application under scavenged and battery power.

We showed in \cite{Guo2022-mlpf-denoising-pami} how a tiny \tip{mlpf} denoises with greater accuracy than prior handcrafted correlation-based denoisers.  
Our \tip{mlpf} runs in batch mode on a desktop \tip{gpu} with a measured throughput of about $10^6\text{ev/s}$\footnote{In batches of 1000 events on NVIDIA RTX 2080 SUPER}, 
but clearly this loses its potential benefit of reducing system-level power because it requires continuous activation of an expensive and power-hungry \tip{gpu}. 
Our work here to realize the \tip{mlpf} within the camera builds on the developments of ultra-quick classifiers from the particle physics community, which require latencies more than 1,000 times quicker (i.e. ns to $\mu$s).

\section{Novel Contributions}
\label{sec:contributions}
\begin{enumerate}
\item Our work reports the first use of ultra-quick latency hardware neural networks developed for particle physics to enable event-by-event accurate event camera denoising.
\item \cref{subsec:network_quantization} shows how to train the \tip{mlpf} so that the signal-noise discrimination accuracy with 4-bit weight and activation precision is nearly as good as the floating point version reported in~\cite{Guo2022-mlpf-denoising-pami}.  \vspace{-0.2cm}
\item Sec.~\ref{sec:implementation} describes the first \tip{fpga} and \tip{asic} implementations of the \tip{mlpf} denoiser. 
\vspace{-0.2cm}
\item Our \tip{asic} implementation of an event camera denoiser is, to our knowledge, the first reported. Our \tip{asic} \tip{mlpf} consumes only a fraction of the power of recently reported event cameras.  \vspace{-0.2cm}
\end{enumerate}
 
\section{Background and Related Work}
\label{sec:background}

Correlation-based \tip{nnb} denoising algorithms,
such as the \tip{baf}~\cite{Delbruck2008-tokyo-frame-free} and the improved \tip{stcf}~\cite{Guo2022-mlpf-denoising-pami} have accuracies that are competitive with complex software algorithms and large neural network denoisers~\cite{Wu2020-prob-graph-denoising,Alkendi2021-gnn-denoiser,Baldwin2020-edncnn,Zhang2021-denoising-with-dnns,Duan2021-eventzoom,Fang2022-aednet-dvs-denoiser}. These correlation denoisers require only a handful of operations per event and have a memory cost about the same as the number of pixels. However, in ~\cite{Guo2022-mlpf-denoising-pami}, we showed that training a tiny \tip{mlp} to discriminate signal versus noise results in a significant improvement in discrimination accuracy, particularly at high noise rates and for more challenging denoising, such as driving, where a moving camera creates a denser structure. 

In \cite{Guo2022-mlpf-denoising-pami}, we constructed datasets of known signal and noise events and measured the denoising accuracy using a \tip{roc} curve and computing the \tip{auc}.

\begin{figure}[tb]
	\centering
\includegraphics[width=.8\linewidth]{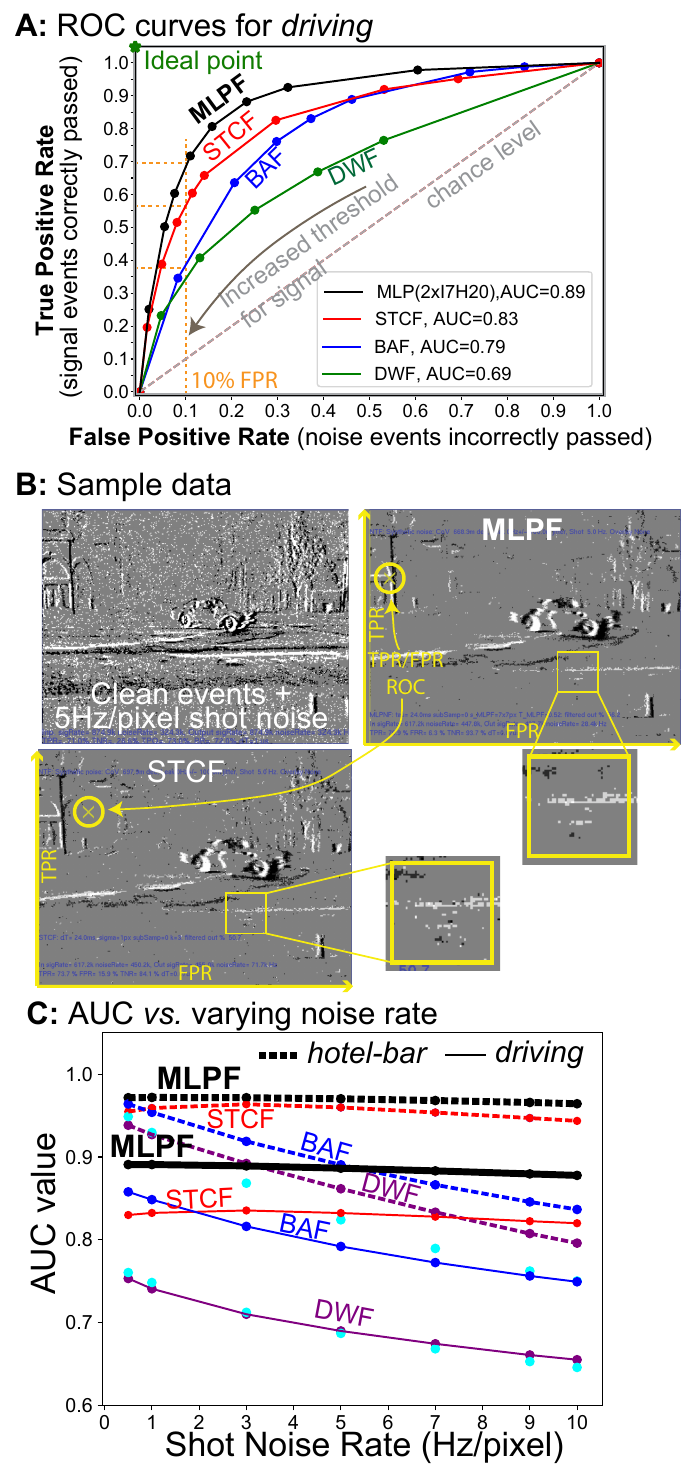}
	\caption{\textbf{A:} \tip{roc} curves for the proposed \tip{mlpf} compared with other denoisers \tip{stcf}, \tip{baf}, and \tip{dwf}~\cite{Guo2022-mlpf-denoising-pami}.
    The signal vs.\ noise discrimination threshold is swept to generate each curve. Evaluated on the \textsl{driving} dataset from \cite{Guo2022-mlpf-denoising-pami} from which this figure is adapted.
    \textbf{B:} Sample data with closeup.
   \textbf{ C: }Dependence of overall accuracy metric \tip{auc} on noise rate.
    }
    \label{fig:MLPF_vs_other_denoisers}
\end{figure}

\Cref{fig:MLPF_vs_other_denoisers} compares the accuracy of the \tip{mlpf} against other low-cost fast denoisers. \Cref{fig:MLPF_vs_other_denoisers}A shows that at \tip{fpr}=0.1, the \tip{tpr} of \tip{mlpf} is approximately 25\% better than the next best \tip{stcf}, and is 2X better than the popular \tip{baf}. 
\Cref{fig:MLPF_vs_other_denoisers}B visually compares the \tip{mlpf} with the \tip{stcf} on the \textsl{driving} dataset with identical \tip{tpr} settings of their thresholds. 
The differences are subtle, but the inset shows how \tip{stcf} allows noise
to pass that \tip{mlpf} blocks.
\Cref{fig:MLPF_vs_other_denoisers}C shows that the \tip{mlpf} maintains \tip{auc} better than other denoisers as the noise rate increases.

Previous implementations of hardware event camera denoising are
\cite{Liu2015-mixed-signal-baf, Khodamoradi2018-order-n-denoiser,Linares-Barranco2019-low-latency-DVS-filtering,Guo2020-hashheat-order-c-denoiser,Kowalczyk2022-iir-filter-denoiser,Barrios-Aviles2018-ldsi-denoiser}\footnote{We exclude denoisers that build a binary image and denoise it\cite{Bose2021-le,Cheng2021-rv} because they discard event timing and event count information.} 
The \tip{fpga} \tip{onf}~\cite{Khodamoradi2018-order-n-denoiser} and HashHeat~\cite{Guo2020-hashheat-order-c-denoiser} require much less memory than the number of pixels that are practical for cameras that only need to denoise scenes with sparse activity regions; 
however, \cite{Guo2022-mlpf-denoising-pami} showed that these denoisers have poor accuracy for dense scenes where the memory is overwritten too quickly 
(see, for example, the DWF curve in \cref{fig:MLPF_vs_other_denoisers}).  The \tip{fpga} \tip{baf} reported in~\cite{Linares-Barranco2019-low-latency-DVS-filtering} --- 
also implemented within some Inivation event cameras --- is useful, but its accuracy and resiliance to high noise rates are much worse than the \tip{mlpf} proposed here; 
see \cref{fig:MLPF_vs_other_denoisers}. The mixed-signal subsampled \tip{baf} of \cite{Liu2015-mixed-signal-baf} does not scale to advanced digital processes. 
The IIR filter array of \cite{Kowalczyk2022-iir-filter-denoiser} 
 severely subsamples the pixel array onto its 2D memory, resulting in significant spatial artifacts.
 The \tip{ldsi} \cite{Barrios-Aviles2018-ldsi-denoiser}
 is a two-layer retina-inspired denoiser that likely has a high latency.

\section{Event camera denoising}
\label{sec:denoising}

\begin{table} 
\centering
\caption{Hardware \tip{mlpf} specifications.}
\label{tab:specs}
\begin{threeparttable}
\renewcommand\TPTtagStyle{\textcolor{red}}
\begin{tabular}{r|c|l}
\hline
$W\times H$  & \tip{dvs} width$\times$height & $346\times 260$\tnote{a}\\
$s_\text{MLPF}^2$  & input patch & ${7^2\text{px}}$ \\
$N_\text{MLPF}$  & \# hidden units     & \nhidden \\
$\tau$ & age window & 64\,ms\tnote{b}\\
\hline
\multicolumn{2}{r|}{Quantization bits} &  (weights/activations)\tnote{c}\\
\cline{2-3}
\multicolumn{2}{r|}{Input units} & 4+1/4+1 \\
\multicolumn{2}{r|}{Hidden units} & 4+1/4 \\
\multicolumn{2}{r|}{Output unit} & 4+1/15+1 \\
\multicolumn{2}{r|}{Threshold $T_\text{MLPF}$} & 15+1 \\
\multicolumn{2}{r|}{Accumulators} & 16.6\tnote{d} \\
\hline
\multicolumn{2}{r|}{Network} & \\
\cline{2-3}
\multicolumn{2}{r|}{Num. weights+bias} & 1001 \\
\multicolumn{2}{r|}{Sparsity} & 40\%\tnote{e} \\
\multicolumn{2}{r|}{Accuracy (\tip{auc})} & 0.87\tnote{f} \\
\end{tabular}
\begin{tablenotes}[para]
\footnotesize 
\item[a] The \tip{tpi} memory has size $W\times H \times{18}$ bits. The 18 bits are composed of a 16-bit ms timestamp and 2-bit polarity.
The ms timestamp is obtaind by right-shifting the $\mu$s timestamp by 10 bits.
\item[b] The age window $\tau$ is quantized power of 2 from 1\,ms to 256\,ms.
\item[c] 4+1 means 4 fraction + 1 sign bit. 4 means unsigned 4-bit fraction.
\item[d] All neurons use default \QKeras signed 16-bit accumulators with 6-bit integer part. The neuron activation function quantizes this activation to produce the output value.
\item[e] Percent of non-zero weights.
\item[f] Evaluated on combined datasets from \cite{Guo2022-mlpf-denoising-pami}.

\end{tablenotes}

\end{threeparttable}
\end{table}

\subsection{Denoising cost metrics }
\label{subsec:cost_metrics}
A practical physical implementation of denoising must consider a quartet of \tips{fom}: discrimination accuracy, silicon area, power, and latency. 
These \tips{fom} generally trade-off against each other, requiring choosing a balance between them. 
High accuracy is needed to remove noise but not signal. A small area is important to minimize cost.
Low latency is needed to discriminate an event before the next one arrives.
Low power is essential for power-constrained applications and to minimize camera heating, 
which increases the photodiode's dark current.

\subsection{Denoising accuracy metrics (ROC and AUC)}
\label{subsec:accuracy_metrics}

As discussed in \cite{Guo2022-mlpf-denoising-pami}, denoising event camera output consists of making a binary discrimination between signal and noise for each event. 
Positive classification means that the event is classified as a signal event. 
The \tip{roc} measurement of discrimination trade-off plots the \tip{tpr} and \tip{fpr} over
all thresholds. 
Ideal denoising achieves zero \tip{fpr} (noise misclassified as signal) and perfect \tip{tpr}=1 (signal correctly classified as signal), resulting in \tip{auc}=1. A higher \tip{auc} means a better classifier.
The optimum \tip{tpr} and \tip{fpr} depend on the application: An always-on surveillance system might favor small \tip{fpr} (large noise suppression), while a mobile robot might favor high \tip{tpr} (high signal retention).
We adopt the metric \tip{auc} used in \cite{Guo2022-mlpf-denoising-pami} to remove bias by a particular choice of threshold.
The \tip{auc} is a scalar measure of the \tip{roc} curve.

\subsection{The MLPF}
\label{subsec:mlpf_algorithm}

The \tip{mlpf} is a lightweight classifier trained on labeled data. It achieves good denoising accuracy by detecting spatiotemporal structural cues that can be helpful in discriminating signal versus noise events. For example, it infers that an event is a likely signal event because it is part of a moving edge or corner.

\begin{figure}[!htb]
    \centering
    \includegraphics[width=\linewidth]{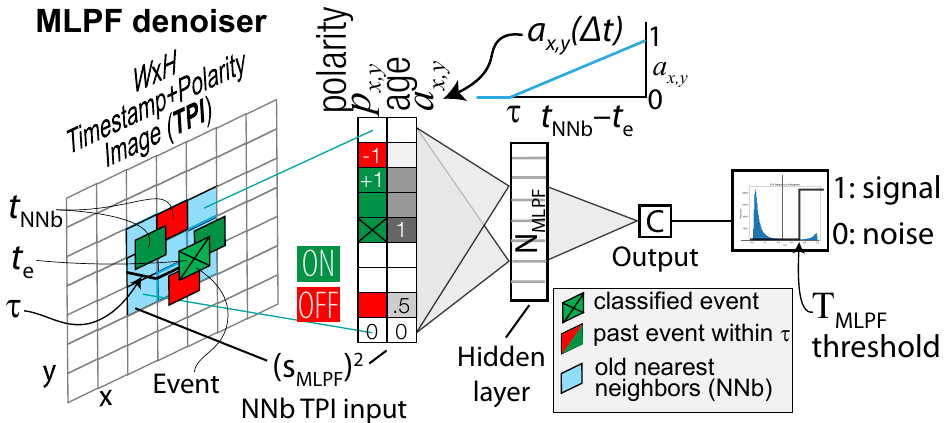}
    \caption{\tip{mlpf} input and network.}
    \label{fig:mlpf-architecture}
\end{figure}

\Cref{fig:mlpf-architecture} illustrates the \tip{mlpf}. It drives a single hidden layer of $N_\text{MLPF}$ neurons from an input patch of $(s_\text{MLPF})^2$ pixels of the \tip{tpi} from the neighborhood around the event that is to be classified. 
The \tip{tpi} is an $W\times H$ pixel 2D memory that holds the latest timestamp (in ms) and the ON or OFF brightness change event polarity corresponding to each DVS pixel. 
\Cref{tab:specs} lists values used in this paper. 


Recent events are important; older events are less relevant, so the $a_\text{x,y}$ input channel encodes the age of \tip{nnb} events as a type of time surface~\cite{Gallego2020-survey-paper}:
$a_{x,y}$ is calculated from each \tip{tpi} pixel by the function illustrated in \cref{fig:mlpf-architecture} where $t_{e}$ is the timestamp of the event \textit{e} that is to be
classified, and $t_\text{NNb}$ is the timestamp of the most recent
previous event stored in the \tip{tpi} corresponding to a \tip{nnb} pixel.
$a_{x,y}$ approaches $1$ for recent events and decays to $0$ for older events.  $\tau$ is the time window parameter.

Polarities of past events are also informative because a moving edge usually produces identical polarities, so the
the $p_\text{x,y}$ input provides the signed \tip{nnb} polarities, using -1 for OFF, +1 for ON and 0 for events older than $\tau$. 
The $p_\text{x,y}$ at the central pixel is from the classified event to provide the necessary information to determine whether the classified event has the same polarity as the past events in the \tip{nnb}.

The \tip{mlpf} with $s_\text{MLPF}=7$ pixels has $7 \times 7 \times 2 = 98$ input units (polarity and age) and a single hidden layer of $N_\text{MLPF}=\nhidden$ neurons, and there are about 1k weights.   
Hidden neurons use a \tip{relu} activation function, and the final output $C$ uses a sigmoid activation function. 
We threshold $C$ against $\text{T}_\text{MLPF}$ to form the final binary discrimination, signaling that the event is signal or noise. 
$\text{T}_\text{MLPF}$ adjusts the \tip{tpr}/\tip{fpr} trade-off.

\section{Implementation}
\label{sec:implementation}

\begin{figure}
    \centering
    \includegraphics[width=\linewidth]{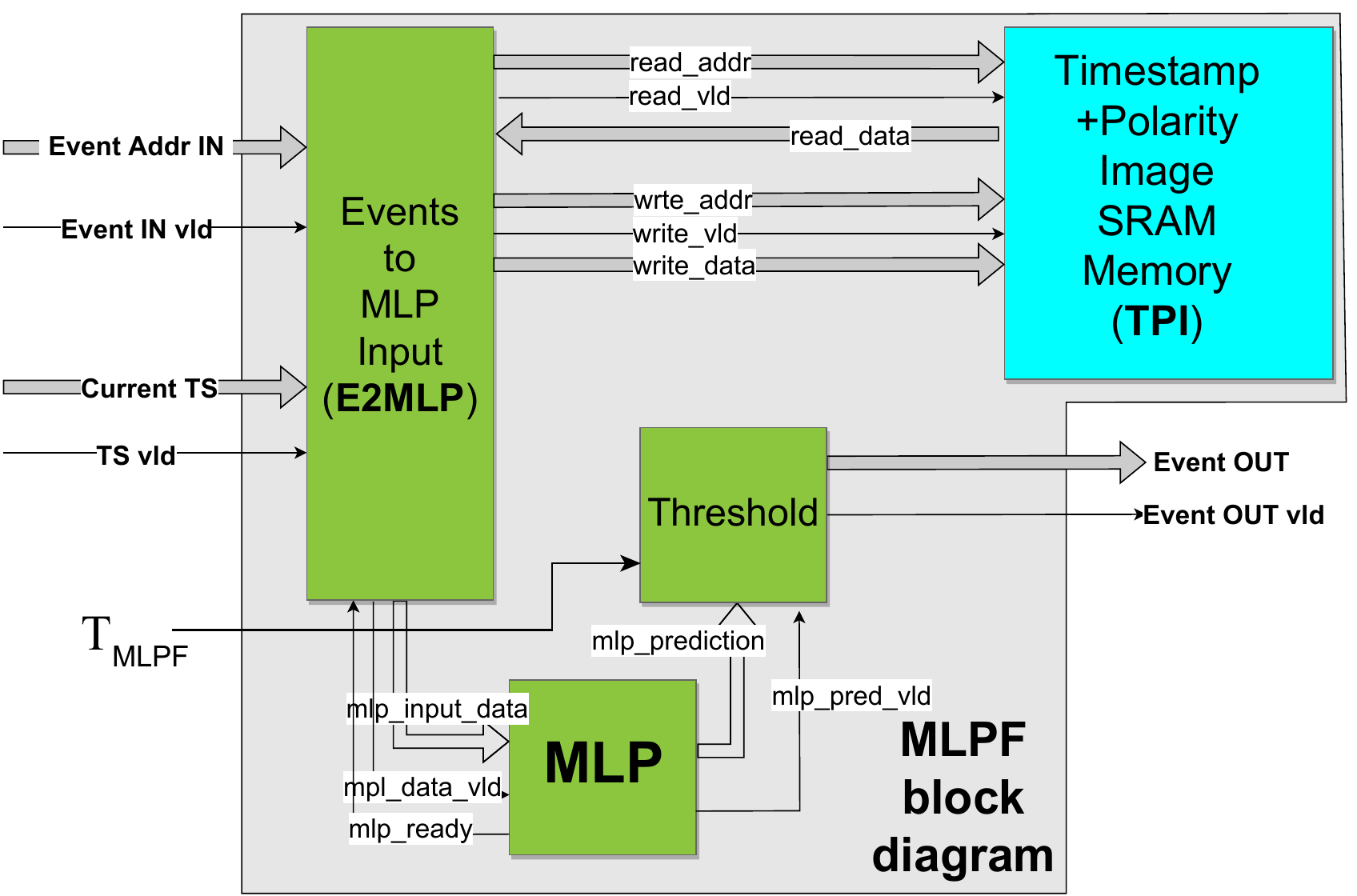}
    \caption{\tip{mlpf} \tip{fpga} and \tip{asic} block diagram.}
    \label{fig:fpga-block-diagram}
\end{figure}

\cref{fig:fpga-block-diagram} illustrates the three major components of the \tip{mlpf} hardware implementation: \tip{e2mlp}, \tip{mlp}, and \tip{tpi}.
The \tip{e2mlp} block receives the event generated by the \tip{dvs} and its timestamp and produces the activation vector for the \tip{mlp}. To calculate this activation vector, the coordinates of the current event are used to access the \tip{tpi} memory and read the timestamps and polarity of the events in the $7 \times 7$ neighborhood of the current event. As the readings are taken, the age of the neighborhood event is calculated from the $a_\text{x,y}$ equation illustrated in \cref{fig:mlpf-architecture}~\cite{Guo2022-mlpf-denoising-pami}. 
The activation vector is composed of 98 elements, where the first 49 are the ages and the last 49 are the polarities of the neighborhood events. After the activation vector is generated, the \tip{tpi} memory is updated with the current event's timestamp and polarity information.

\subsection{Dataset, Training, and Network quantization}
\label{subsec:network_quantization}
Our training dataset combined the \textsl{hotel-bar} and \textsl{driving} noise datasets from~\cite{Guo2022-mlpf-denoising-pami}. 
These datasets were chosen to cover the range of applications from sparse surveillance to dense mobile robotics event camera output.
To these clean \tip{dvs} recordings we add both simulated and pre-recorded \tip{dvs} \tip{ba} noise events.
We train the network to optimally discriminate signal and noise events.
Appendix F of the supplementary material of 
 \cite{Guo2022-mlpf-denoising-pami} details the floating point \tip{mlpf} training procedure (we removed the dropout layer here). 
 We used exactly the same dataset and training procedure except for quantization-aware training
 as detailed next. 

Applying model compression at training time is crucial to minimize the area and maximize the accuracy of the model. We rely on quantization-aware training through the \QKeras\footnote{\url{https://github.com/google/qkeras}.} library.
Through drop-in replacement of standard Keras layers, this library allows users to create quantized versions of the Keras equivalent. Doing so at training time optimizes accuracy by allowing the network to adjust the model weights according to the precision tailored for the hardware.

\begin{figure}
    \centering
    \includegraphics[width=\linewidth]{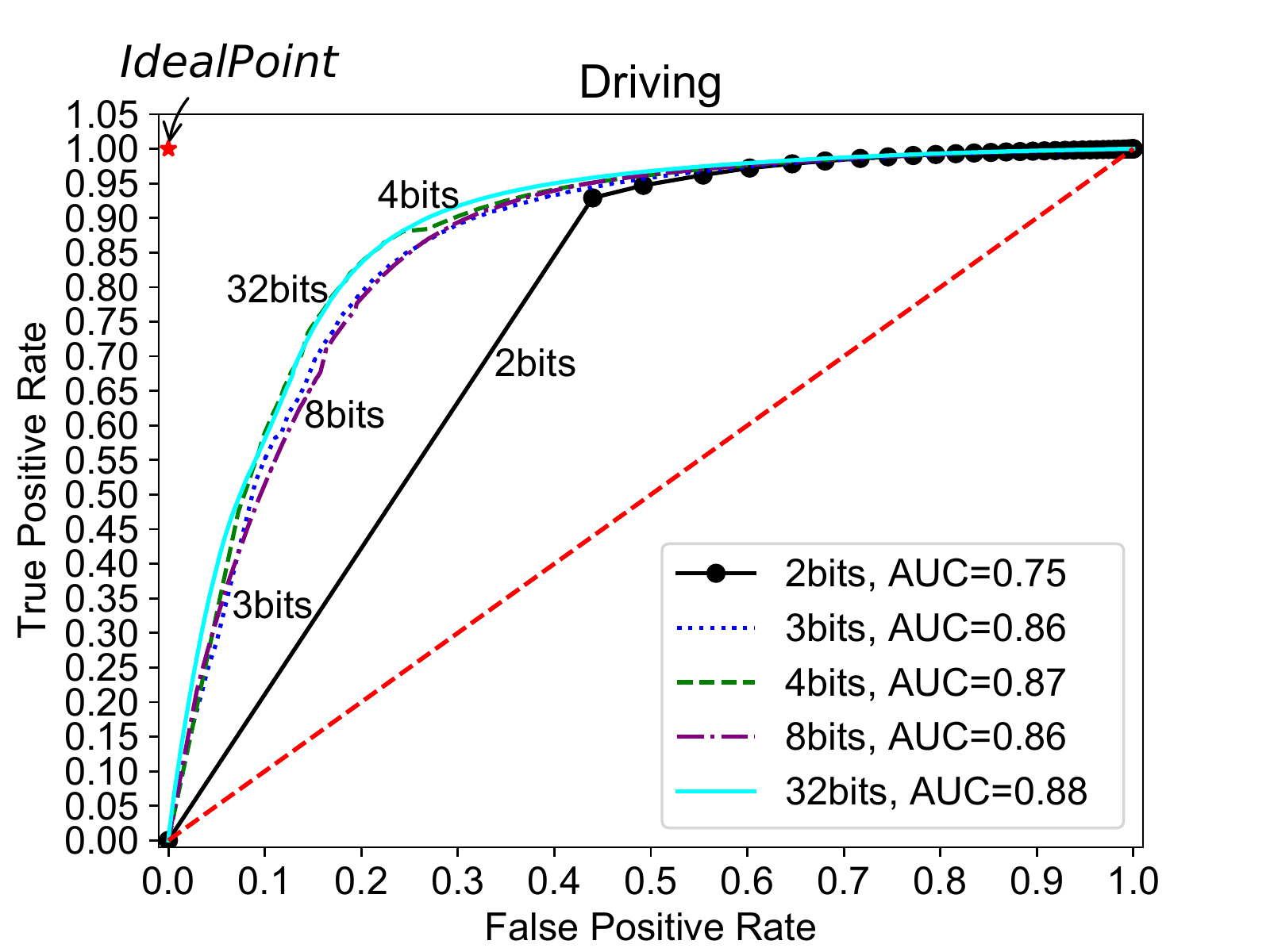}
    \caption{\tip{roc} curves for quantized \tips{mlpf} compared with floating point 32bits network. Evaluated on the more challenging \textsl{driving} dataset from \cite{Guo2022-mlpf-denoising-pami}. }
    \label{fig:quantization_roc}
\end{figure}

\Cref{fig:network_code} lists the network code. The quantization function \texttt{quantized\_bits} takes as input the number of bits to quantize to, the number of integer bits, as well as an optional parameter \textit{alpha}, which is used to change the absolute scale of the weights while keeping them quantized within the same bit width. In our case, we quantize the kernel and bias to a bit width of 4 with zero integer bits and set \textit{alpha} to 1 (meaning that no scaling is applied).
The inputs, weights, activations, and output are quantized to a fixed point decimal ranging from -1 to 1\footnote{Actually to the closest possible value to 1, e.g. 0.875 for 4-bit signed-value quantization}.
Note that the function \texttt{quantized\_bits} behaves differently from the fixed point numbers used in Vivado \tip{hls}. For \texttt{quantized\_bits(bitwidth, integer bits)}, the HLS equivalent is \texttt{ap\_fixed(bitwidth, integer bits+1)}.
The experimental results in \cref{fig:quantization_roc} show that the use of quantified weights and activations has almost no effect on the \tip{auc} value, except for 2 bits. Due to the limited diversity of the \tip{mlpf} output caused by the 2-bit input and weights of the last dense layer, sweeping thresholds barely affect the \tip{roc} curve and the \tip{auc}.

For our implementation, we chose 4-bit weight and activation quantization. 
For simplicity of design, the input of the network is quantized to 4 bits for age $a$ and polarity $p$ channels.
The age is computed by right shifting the $\Delta t$ time difference by some number of bits according to 
the desired $\tau$. 
The polarity is represented as 0 or $\pm 1$.
The final output is quantized to 16 bits for a fine discrimination threshold.
The chosen model (\cref{fig:network_code}) achieves \tip{auc} of 0.87 on the \textsl{driving} dataset, close to the floating point \tip{auc} of 0.88 (\cref{fig:quantization_roc}).
%

\lstset{
language=Python,
basicstyle=\footnotesize\ttfamily,breaklines=true,tabsize=2
} 

\begin{figure}[btp]
    \caption{QKeras code for the network.}
    \label{fig:network_code}
\begin{lstlisting}
inputs = Input(shape=[98, ], name='input')
x = QActivation("quantized_bits(4,0,alpha=1)", name="qact0")(inputs)
x = QDense(10, input_shape=(98, ), kernel_quantizer=quantized_bits(4,0,alpha=1),bias_quantizer=quantized_bits(4,0,alpha=1),name="fc1")(x)  
x = QActivation("quantized_relu(4,0)", name="relu1")(x)
x = QDense(1, kernel_quantizer=quantized_bits(4,0,alpha=1),bias_quantizer=quantized_bits(4,0,alpha=1),name="fc2")(x)
x = Activation("sigmoid", name="doutput")(x)
model = Model(inputs, x)   
\end{lstlisting}
\end{figure}





\subsection{Firmware implementation with \hlsfml}
\label{subsec:hls4ml}

We use the \hlsfml library~\cite{Duarte:2018ite} to generate an FPGA implementation of the \tip{mlp}. \hlsfml converts a given \tip{dnn} into \tip{hls} code, which uses a C-like syntax decorated with hardware-specific optimizations such as pipelining and custom data types.  The \tip{hls} compiler then translates the \tip{hls} code into the \tip{fpga} firmware. In contrast to conventional \tip{rtl} logic design, \tip{hls} lies at a higher level of abstraction, making it faster to specify and generate custom FPGA architectures, since it is a subset of C coding and thus allows greater productivity, but it has limitations for designs that require complex interconnected circuits.

The \hlsfml library interfaces to \QKeras~\cite{qkeras_hls4ml} to provide a translation of the quantized weights of the \tip{mlpf} model using quantization-aware training into the correct fixed-point equivalent values on the FPGA implementation. 
A developer can compile the \tip{hls} code for the network in a few seconds from the trained \QKeras model.

\hlsfml was developed for low-latency and high-throughput applications. Thus, \hlsfml generates a custom implementation for each layer and the entire network is spatially instantiated on the \tip{fpga} logic. Our \tip{mlpf} is implemented in a fully parallel manner; all multiplications are performed in parallel. The output neuron activation is computed as a combinational logic weighted sum of the hidden units, which themselves are computed as combinational logic weighted sums of input unit activations.
The weights are frozen into the connections, and the network must be resynthesized when they need to be changed. Zero weights do not use hardware resources.

The Vivado \tip{hls} compiler
limits arrays to 1024 elements, which is less than what the hidden layer of the 98I-20H-1O \tip{mlpf} in \cite{Guo2022-mlpf-denoising-pami} would require to be fully unrolled\footnote{The array would have to be partitioned into 98*20=1,960 elements, which is more than the tool allows.}.
Therefore, to reach the lowest possible latency, we trained an \tip{mlpf} with $N_\text{MLPF}=10$ hidden units.
Using 10 hidden units instead of 20 and the chosen quantization (\cref{tab:specs}) reduces the \tip{auc} from the floating point 0.89 to 0.87.
The low precision of the weights allows all multiplications in the network to be computed by \tips{lut} rather than the more power-hungry hardware multiplier units. 


Since we do not depend on having an output between zero and one (this is only necessary at training time), we removed this final sigmoid activation function and instead placed a threshold on the activation of the output unit. 
This saves one clock cycle. 



\subsection{FPGA realization}
\label{sec:fpga_implementation}

\begin{figure}
    \centering
    \includegraphics[width=0.7\linewidth]{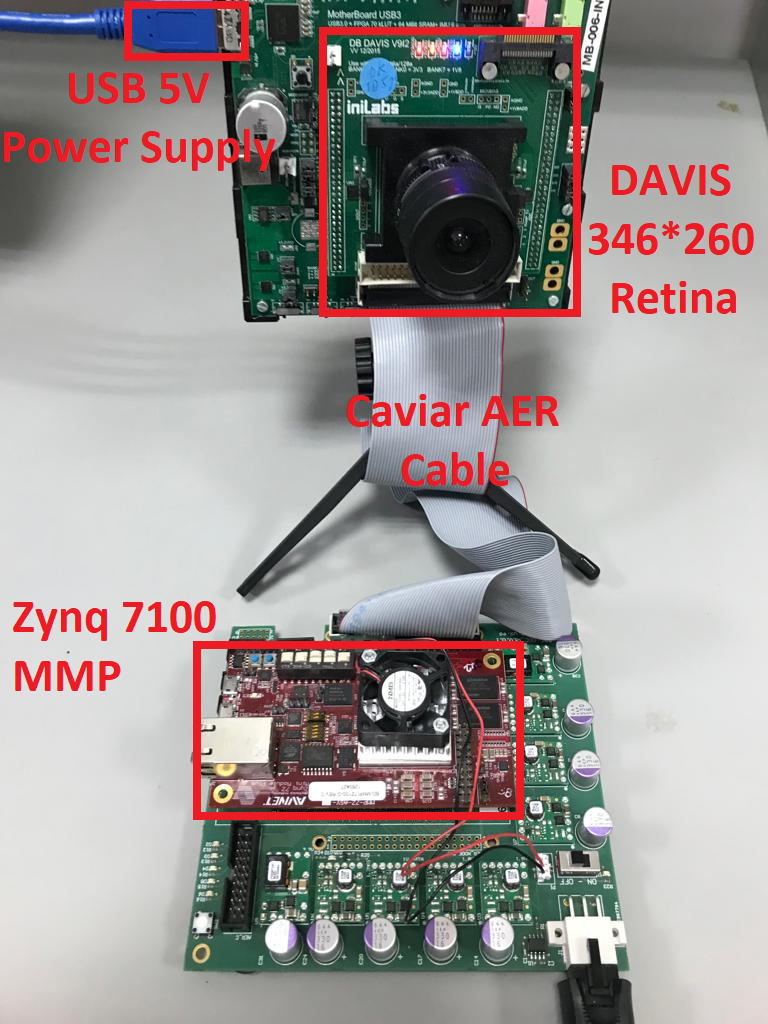}
    \caption{The DAVIS camera connected to the Zynq XC7Z100 \tip{fpga} development board that holds the \tip{mlpf}.}
    \label{fig:davis_fpga_photo}
\end{figure}

\cref{fig:davis_fpga_photo} shows our setup to implement \tip{mlpf} using a DAVIS346 prototype camera~\cite{Brandli2014-davis,Taverni2018-bsi-vs-fsi-davis} connected by \tip{aer} cable to a custom Xilinx xc7z100 \tip{fpga} development board~\cite{Linares-Barranco-icons21-zynq7100-npp}.

Our first implementation of the \tip{e2mlp} module used dual read port \tip{bram} to read the 49 \tip{tpi} pixels in 25 clock cycles.
By adopting a memory organized as in Liu's EDFLOW camera~\cite{Liu2022-edflow}, the columns are read in parallel, in only 7 clock cycles. 
The key to making this scheme work was the correct use of the Vivado \tip{hls} pragma
 \textsl{array\_partition}  as in \cite{Liu2022-edflow}. This pragma partitions the array into smaller arrays; in this case, a separate \tip{bram} is for each sensor column. The resulting \tip{rtl} has multiple smaller memories instead of one large memory. This pragma effectively increases the amount of read and write ports for storage. The resulting \tip{bram} usage increases because it requires more memory instances and registers.

The activations generated by the \tip{e2mlp} module are used by the \tip{mlp} module to perform inference on these data and generate a prediction. To decide whether the current event is considered to signal or noise, a threshold is applied to the calculated prediction.

For the \tip{fpga} implementations, the \tip{e2mlp}, \tip{mlp} and \tip{tpi} blocks are written in Vivado \tip{hls} 2020.1. 
We synthesized our solution for two different \tips{fpga}, a decade-old midrange Xilinx Zynq XC7Z100 and an entry-level Zynq Ultrascale+ ZU3CG. 

Both of these \tip{soc} \tips{fpga} are attractive candidates for building an event camera because their processors can run a simplified Linux kernel and thus the camera could form a discrete smart camera node in a robot.

\subsection{ASIC realization}
\label{sec:asic_implementation}

\begin{figure}
    \centering
    \includegraphics[width=0.6\linewidth]{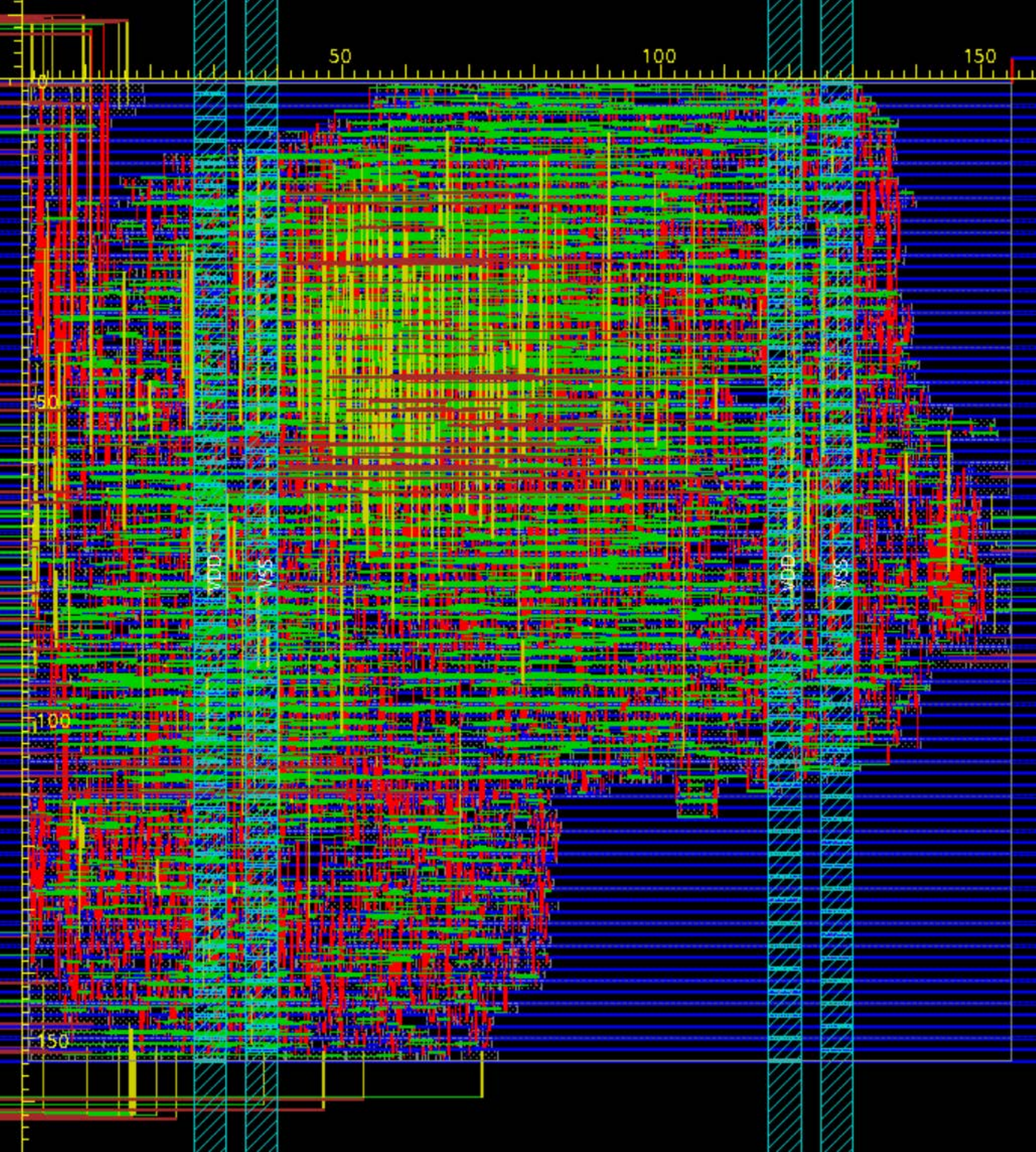}
    \caption{ASIC \tip{mlpf} \tip{e2mlp}+\tip{mlp} layout in 65nm digital technology. 
    The ruler units are microns. The logic area is 0.022mm$^2$ (67\% density). The bonding pads are not included since this block would normally be placed adjacent to the sensor core.
    In our current version, the \tip{sram} blocks are not included.}
    \label{fig:asic-layout}
\end{figure}

We wrote an \tip{rtl} (SystemVerilog) implementation of the \tip{mlp} and \tip{e2mlp} blocks in \cref{fig:fpga-block-diagram} to circumvent current Vivado \tip{hls} limitations in generating \tip{asic} compatible \tip{rtl}. We target a conventional 65nm digital process technology from TSMC, since it is economical, widely available, and because we had access to it for this project. We used Cadence Genus for the synthesis and Cadence Innovus for the place and route. \cref{fig:asic-layout} shows the layout of the circuit without memory. It was obtained using the low power library, applying the clock-gating technique (for 43\% of inferred flip-flops) and reducing the power ring size to minimize its area. 
Our \tip{mlp} architecture also allows us to reuse the multipliers, adders, and registers, minimizing the chip area and power consumption. 
We use data gating to prevent the gates and flip-flops from switching on clock cycles without new data.
We measured an improvement of 20\% in power consumption with the use of clock gating. Our \tip{rtl} design and scripts are released as open source under a GPL license.

Our \tip{asic} architecture differs from the \tip{fpga} implementation to address the following trade-offs. The \tip{fpga} implementation uses one \tip{bram} per column of the image, to be able to read any arbitrary $7\times 7$ neighborhood within 7 clock cycles. This is possible since these \tips{sram} are synthesized as groups of much smaller \tips{bram} already available on the \tip{fpga}. Since \tips{sram} are custom generated on silicon, similar \tip{asic} implementation with one \tip{sram} per column would result in a much higher area and consume more power due to the duplication of control circuitry for each \tip{sram}. 

Therefore, in our \tip{asic} design, we utilize two 1W1R \tips{sram}\footnote{Our memory compiler could not generate dual-read-port \tip{sram}, so we used two single port \tips{sram}}. We can read two pixels of the \tip{tpi} per cycle, compared to seven per cycle in our \tip{fpga} design. As a result, the latency of the \tip{asic} implementation is 33 clock cycles, compared to 10 clock cycles in \tip{fpga}. Since our \tip{asic} implementation has a much higher clock frequency, it is faster with a smaller latency (40 ns) compared to the \tip{fpga} (43 ns).

\section{Results}
\label{sec:results}
\glsunset{ff}
\glsunset{lut}
\glsunset{bram}
\begin{table*}
\centering
\begin{threeparttable}[!t]
\renewcommand\TPTtagStyle{\textcolor{red}}
\caption{\tips{fom} of the hardware \tips{mlpf} for \cref{tab:specs} parameters.}
\label{tab:results}

    \begin{tabular}{c|cc|cc|cccc}
    & \multicolumn{2}{c|}{\underline{Xlinx Zynq XC7Z100}} &  \multicolumn{2}{c|}{\underline{Xlinx Zynq U+ ZU3CG}} & \multicolumn{4}{c}{\underline{65nm \tip{asic}. Vdd=1.08V}} \\
       Max. Clock Freq.  & \multicolumn{2}{c|}{100\,MHz} &  \multicolumn{2}{c|}{236\,MHz} & \multicolumn{2}{c}{833\,MHz} &&  \\
       Latency & cycles & ns & cycles & ns & cycles & ns &\\
       \tip{e2mlp}  & 7 & 70 & 7 & $\sim$30 & 30 & 36\\
       \tip{mlp}  & 3 & 30 & 3 & $\sim$13  & 3 & 3.6\\
       Total  & 10 & 100 & 10 & $\sim$43 & 33 & 40 \\ 
       \hline
    \end{tabular}
    
   \begin{tabular}{c|ccc|ccc|cc}
    & \multicolumn{3}{c|}{\underline{Xlinx Zynq xc7z100}} &  \multicolumn{3}{c|}{\underline{Xlinx Zynq U+ ZU3CG}} & \multicolumn{2}{c}{\underline{65nm \tip{asic}}} \\
       & \tip{lut}  & \tip{ff} & \tip{bram} & \tip{lut}  & \tip{ff} & \tip{bram} & Logic area & \tip{sram} area\\
    Resources & 18.4k & 3.9k & 400\tnote{a}  &  24.6k & 3.8k & 400\tnote{a} & 0.022mm$^2$ & 4.3\tnote{b} mm$^2$ \\
     Resource \% &  6\% & 0.72\%  & 26\% & 34\% & 2\% & 92\% & -- & -- \\
     \hline
     Power & \multicolumn{6}{c|}{Not relevant} & \multicolumn{2}{c}{40mW\tnote{c}}\\
       Energy/event  & \multicolumn{6}{c|}{Not relevant} & \multicolumn{2}{c}{4nJ\tnote{d}} \\
    \hline
    \end{tabular}
\begin{tablenotes}[para]
\footnotesize 
\item[a] Used for \tip{tpi}.
\item[b] Our two \tips{sram} each run at 1\,GHz with 1.0V supply. They have 18 bit word width and 2048 words per bank. Maximum possible is 2048 per bank using multiple banks to store the \tip{tpi}. The entire design requires $W\times H /2048 \approx 44$ such \tip{sram} banks.
\item[c] Assumes 1\,MHz event rate and 4\,nJ per event (1.2\,nJ \tip{mlp} + 2.4\,nJ \tip{sram} +0.4\,nJ \tip{e2mlp}), plus 35\,mW \tip{sram} leakage.
\item[d] Assumes one \tip{mlp} inference and 50 \tip{sram} accesses. The energy per event is computed by $E=P\times T\times N$, where $P$ is the power consumption of the block, $T$ is the period of the clock and $N$ is the required number of clock cycles per event. 
\end{tablenotes}
\end{threeparttable}
\end{table*}

\glsreset{ff}
\glsreset{lut}
\glsreset{bram}
\cref{tab:results} lists the main \tips{fom} of the \tip{fpga} and \tip{asic} implementations of the \tip{mlpf} in terms of maximum clock frequency, latency in clock cycles and nanoseconds, resources used, power, and energy per event.
\tip{fpga} resources consist of \tips{lut}, \tips{ff}, and \tips{bram}.
A \tip{lut} is a programmable memory element (at synthesis time) that implements combinational logic. A \tip{ff} is a single bit memory element.
\tip{asic} resources are measured by area.
We did not measure \tip{fpga} power consumption because it is completely dominated by standby power.
The \cref{tab:results} notes provide details for many of the specifications.

The \tip{mlp} inference latency is only 3 cycles, giving a total \tip{fpga} latency of 10 cycles.
In our \tip{fpga} implementations, synthesis generated the \tip{tpi} as multiple dual port \tip{bram} memories and needs 400 BRAM\_18K. This memory has 89960 words ($W\times H$ pixels) with 18-bit word size, where 16 bits are used to store the timestamp and 2 bits for the event polarity.

Our \tip{asic} implementation area is dominated by \tip{sram}. We estimated a standby power of about 35\,mW, consisting mostly of \tip{sram} leakage. The energy per event is 4\,nJ, so a noise event rate of 1\,MHz would burn another 4\,mW.


\section{Conclusion}
\label{sec:conclusion}

This paper proposes two logic circuit implementations of a multilayer perceptron event camera denoiser.

The \tip{fpga} implementation discriminates each event in 10 clock cycles or 100\,ns with the clock frequency of 100\,MHz, allowing the denoising to occur at event rates of 10\,MHz. 
For higher noise rates, the denoiser can bypass or block events that arrive while it is busy.

We showed two implementations mapped to \tip{soc} \tips{fpga} that would be practical for integrating \tip{mlpf} denoising within an embedded \tip{dvs} camera. Using an individual \tip{bram} for each column of the pixel array, we showed how the \tip{mlp} input can be formed for each event in only 7 clock cycles, enabling 10-cycle discrimination.
The \tip{asic} block, which could be integrated into future \tip{dvs} chips, has a longer latency in clock cycles, but since the block runs at a much higher clock frequency, it can classify each event in 40\,ns, enabling event-by-event denoising up to rates of 25\,MHz. This \tip{asic} block power consumption is estimated to be on the order of 40\,mW, which is a fraction of the power of recently reported event cameras.%
\footnote{Recent event camera chips ~\cite{Guo2023-omnivision-isscc-hybrid,Kodama2023-sony-isscc-hybrid-1p22um,Niwa2023-sony-isscc-3um-hybrid-davis} report sensor power consumption from 60\,mW to 525\,mW.}
At low noise rates, the power consumption drops to essentially the \tip{sram} leakage. 
Our current design uses standard \tip{sram}; by
exchanging this memory for lower leakage
low power memory, we believe it would be possible to reduce
the standby power by an order of magnitude.

\cref{tab:comparison} compares the hardware \tip{mlpf} with other \tip{fpga} hardware denoisers in terms of accuracy, memory cost, and throughput (there is no other \tip{asic} implementation to compare with). 
As discussed in \cref{sec:background}, except for the \tip{baf} in \cite{Linares-Barranco2019-low-latency-DVS-filtering}, 
other denoisers seek to minimize memory, which in general leads to poor discrimination accuracy at high noise rates. 
The \tip{mlpf} has by far the best \tip{auc} accuracy on the sparse \textsl{hotel-bar} and dense \textsl{driving} datasets of \cite{Guo2022-mlpf-denoising-pami}. Although \tip{mlpf} is not the fastest, its maximum denoising rate of $\approx 25$\,MHz is suitable for recent event cameras.

\begin{table}[tb]
 \centering
 \caption{Comparison with other hardware \tip{dvs} denoisers.}
  \label{tab:comparison}%
 \resizebox{\columnwidth}{!}{
 \begin{threeparttable}[htbp]
  \setlength{\tabcolsep}{1.1mm}{
  \begin{tabular}{lcccc}
  \toprule
\textbf{Denoiser}   & \thead{\textsl{hotel-bar} \\\tip{auc}\tnote{a}} & \thead{\textsl{driving} \\ \tip{auc}\tnote{a}} & \thead{Mem(\#)\tnote{b}} & \thead{Max. Event\\ Rate\tnote{c} \\{MHz}}  \\ 
\midrule
\textbf{\tip{mlpf} \tip{fpga}}     &  \textbf{0.96}     &    \textbf{0.87}    & $W\times H$       & 23    \\ 
\textbf{\tip{mlpf} \tip{asic}}  &   ''    &  ''     & $W\times H$       & 25    \\ 
\tip{baf}~\cite{Linares-Barranco2019-low-latency-DVS-filtering} &      0.89 &    0.79   &   $W\times H$     &  3.6 \\
\tipshort{onf}~\cite{Khodamoradi2018-order-n-denoiser}  & 0.01\tnote{e} & 0.01\tnote{e} & $2\times(W+H)$  & 3 \\
HashHeat~\cite{Guo2020-hashheat-order-c-denoiser}   & 0.67 & 0.56\tnote{f} &  \textbf{128}  &  100   \\
IIRs~\cite{Kowalczyk2022-iir-filter-denoiser}   & NA & NA & $W\times H$\tnote{g}   & \textbf{385}    \\
\tip{ldsi}~\cite{Barrios-Aviles2018-ldsi-denoiser}   & NA & NA & $W\times H$   & 3?\tnote{d}    \\
\bottomrule 
\end{tabular}
    }%
  \begin{tablenotes}[para]
  \item[a] Evaluated at shot noise rate of 5\,Hz/pixel using method in \cite{Guo2022-mlpf-denoising-pami}.
\item[b] Memory cells for $W\times H$ pixel DVS assuming cells each have on the order of 18-36 bits and that no subsampling as in \cite{Delbruck2008-tokyo-frame-free,Liu2015-mixed-signal-baf,Kowalczyk2022-iir-filter-denoiser} is used to reduce memory at the cost of reduced denoising accuracy.
\item[c] Maximum event rate for denoising all events.
\item[d] The \tip{ldsi} was measured at 50\,MHz clock and reported to run at up to 177\,MHz clock frequency. The throughput was not reported except to state it could run up to the maximum event rate of \cite{Lichtsteiner2008-dvs} which is 1\,MHz. 
\item[e] The \tip{onf} \tip{auc} is very low because it cannot achieve large \tip{tpr} and \tip{fpr} even with very low threshold.
\item[f] The HashHeat \tip{auc} is derived from sweeping its threshold compared with the heat values stored in a 128-element vector. 
\item[g] The IIRs denoiser actually subsampled the array to 20x20 pixel memory, causing significant artifacts.
\end{tablenotes}
\end{threeparttable}
} 
\end{table}


Recent industrial event camera chips~\cite{Guo2023-omnivision-isscc-hybrid,Kodama2023-sony-isscc-hybrid-1p22um,Niwa2023-sony-isscc-3um-hybrid-davis}
feature event pixels that are less than 5\,$\mu$m and are built in stacked technologies with an optical top part using 90nm technology and a digital bottom part using feature sizes down to 22nm technology. 
The digital layer area is several times larger than the sensor area. 
For this paper, we designed the \tip{mlpf} \tip{asic} block for 65nm technology; In 22nm technology, the logic area would be smaller, but the \tip{sram} required to hold \tip{tpi} would need to be several Mpx. 
The dominant \tip{tpi} \tip{sram} would occupy an area of about 1\,mm$^2$/Mpx\footnote{Based on 6T cell with area $130F^2$, where $F=22\text{nm}$ is the feature size. 
The total \tip{sram} area would be $130\times (\text{22e-9m})^2 \times 1e6\text{px} \times 18\text{bits} \times 1e6\text{mm}^2/\text{m}^2=1\text{mm}^2$.}.
It would likely be possible to use subsampling onto a smaller-resolution \tip{tpi}~\cite{Liu2015-mixed-signal-baf} too much loss of denoising accuracy.
The \tip{mlpf} would comprise only a small part of these designs. 
However, since these recent event camera chips use some form of sampled, frame-like sparse readout of the brightness change events, it is unclear to what extent they can use fine event timing for denoising like \tip{mlpf}.



\subsection*{Acknowledgements}
{\footnotesize
This work was supported by the Swiss SNF projects SCIDVS (2O0O21\_185069/1), VIPS (2082-l\_181010/1), and PZ00P2\_201594, and by the Spanish grant MINDROB (PID2019-105556GB-C33/AEI/10.13039/501100011033).
We thank Dr. Min Liu and Prof. Shih-Chii Liu (INI), Prof. Mingu Kang (UCSD), and anonymous reviewers for their suggestions to improve the paper.}

\clearpage
\AtNextBibliography{\footnotesize}
\printbibliography

\end{document}